\documentclass{article}

     \usepackage[preprint]{neurips_2020}

\usepackage[utf8]{inputenc} 
\usepackage[T1]{fontenc}    
\usepackage{hyperref}       
\usepackage{url}            
\usepackage{booktabs}       
\usepackage{amsfonts}       
\usepackage{nicefrac}       
\usepackage{microtype}      
\usepackage{multirow}
\usepackage{subcaption}
\usepackage{graphicx}
\usepackage{amsmath}
\usepackage{algorithm}
\usepackage{algorithmic}
\usepackage{comment}
\usepackage{tabularx}
    \newcolumntype{L}{>{\raggedright\arraybackslash}X}

\title{Training Large Neural Networks with Constant Memory using a New Execution Algorithm}

\author{%
  Bharadwaj Pudipeddi \\
  Microsoft\\
  Sunnyvale, CA \\
  \texttt{bharadwaj.pudipeddi@microsoft.com} \\
  \And
  Maral Mesmakhosroshahi \\
  Microsoft \\
  Sunnyvale, CA \\
  \texttt{maral.mesmakhosroshahi@microsoft.com} \\
  \AND
  Jinwen Xi  \\
  Microsoft \\
  Sunnyvale, CA \\
  \texttt{Jinwen.Xi@microsoft.com} \\
  \And
  Sujeeth Bharadwaj \\
  Microsoft \\
  Sunnyvale, CA \\
  \texttt{sujeeth.bharadwaj@microsoft.com} \\
}

\begin{document}

\maketitle

\begin{abstract}

Widely popular transformer-based NLP models such as BERT and Turing-NLG have enormous capacity trending to billions of parameters. Current execution methods demand brute-force resources such as HBM devices and high speed interconnectivity for data parallelism. In this paper, we introduce a new relay-style execution technique called L2L (layer-to-layer) where at any given moment, the device memory is primarily populated only with the executing layer(s)'s footprint. The model resides in the DRAM memory attached to either a CPU or an FPGA as an entity we call eager param-server (EPS). To overcome the bandwidth issues of shuttling parameters to and from EPS, the model is executed a layer at a time across many micro-batches instead of the conventional method of minibatches over whole model. L2L is implemented using $16GB$ V100 devices for BERT-Large running it with a device batch size of up to $256$. Our results show $45\%$ reduction in memory and $40\%$ increase in the throughput compared to the state-of-the-art baseline. L2L is also able to fit models up to $50 Billion$ parameters on a machine with a single $16GB$ V100 and a CPU with $512GB$ memory and without requiring any model partitioning. L2L scales to arbitrary depth allowing researchers to develop on affordable devices which is a big step toward democratizing AI. By running the optimizer in the host EPS, we show a new form of mixed precision for faster throughput and convergence. In addition, the EPS enables dynamic neural architecture approaches by varying layers across iterations. Finally, we also propose and demonstrate a constant memory variation of L2L and we propose future enhancements. This work has been performed on GPUs first, but also targeted towards all high TFLOPS/Watt accelerators.

\end{abstract}

\section{Introduction}

The transformer architecture spawned the "ResNet" moment in natural language processing (NLP), where residual blocks of arbitrary depth can be stacked to create state-of-the-art models such as BERT~\citep{bert}, GPT-2~\citep{gpt} and the recently published gigantic GPT-3 with 175B parameters~\citep{gpt3}. Although these models reduce design complexity, they have significant overhead in memory requirements. BERT-large can barely train on a high-end GPU such as the V100 with $16GB$ with a batch-size of $2$.

Training large NLP models like BERT with billions of parameters has only been successfully carried out on high-bandwidth memory devices such as GPUs and TPUs with high memory capacities. The memory size is influenced not only by the model parameters but also by a sufficiently large batch size required for convergence. The transformer-class of models such as BERT can be classified as having high weight/activation ratios: they have high number of parameters and yet relatively small output activations. For instance, BERT-large has $24$ encoder layers, $350M$ parameters, but the layer output size is only $1MB$ per sample. This is the key observation to develop a more efficient execution method for large NLP models.

\subsection{Related Work}
Traditional distributed training of neural networks started with data parallelism which keeps a copy of the whole model on each device and partitions data among multiple devices~\citep{dataparallel1, dataparallel2, dataparallel3, dataparallel4, dataparallel5}. Data parallelism works with the assumption that the whole model can fit on the device which is not necessarily true anymore. Training larger models has been requiring the model to be partitioned across multiple devices~\citep{dataparallel1} using model parallelism approaches which could often be inefficient and hard to implement. Pipelining is another traditional approach that is common for distributed training which overlaps computations between the layers. 

There are more novel approaches proposed in recent years to solve the memory limitation problem. The first is PipeDream~\citep{pipedream}, which partitions a model across multiple devices and pipelines the execution of forward passes interspersing them with backward passes to maximize hardware utilization. Pipedream updates on every minibatch and circumvents staleness by maintaining various versions of the model. A related model parallelism approach is GPipe~\citep{gpipe} which also partitions the model across multiple devices. However, GPipe pipelines the execution of microbatches before applying a single synchronous gradient update for the entire minibatch. GPipe stacks the forward pass output activations and recomputes them during backward pass as it pops each microbatch off the stack. GPipe and PipeDream both have overheads related during the start of the pipeline, and both approaches require the number of devices to scale with the model depth and not just the layer size. Therefore, are not constant memory approaches. Also, neither approach has made specific extensions for distributed data parallelism training over model parallelism that can overcome their overheads.

A third method is OpenAI’s gradient checkpointing~\citep{openai, openai2}. which tradeoffs memory with more computation. A deep neural network can checkpoint a subset of nodes in the computational graph so that it does not need to retain state of all the nodes. For a node’s backward pass, the required activations are recomputed from the nearest checkpoint. A constant memory implementation gradient checkpointing is feasible, but results in a computational complexity that scales by $O(N^2)$ and it's recomputation costs for large models are massive.

vDNN~\citep{vdnn} is another technique proposed by Nvidia where special CUDA memory management techniques convert a model to a layerwise execution on the device where older layers are released to CPU memory based on a layer distance heuristic. vDNN was demonstrated on vision models, and works exceptionally well on a Titan X GPU. However, vDNN is different by nature as the heuristic trades off performance with a coarse grain of layer-level buffering. This will not adapt well to large transformer-based models where the buffering requirements would be high due to the enormous size of the layer and would still cause computational efficiency issues due to smaller memory available for execution. vDNN also was constructed to off-load entire activations to avoid recompute, but even if it only off-loaded output activtions, its method of heuristic choice by CUDA level software based on layer distance instead of carefully orchestrated transfers like L2L cannot hide the transfer latencies or the data parallelism scaling overhead incurred.

The recently published DeepSpeed and Zero\citep{zero} partition a single copy of the model across many GPUs while running them in data parallelism layer-by-layer. DeepSpeed is an effective method for large models as they demonstrate a $17B$ parameters model over $256$ GPUs. But DeepSpeed requires the model to fit across the combined memory of all the GPU devices.

There is no known solution, however, where a large size transformer-based model of billions of  parameters can be run on a single device with insufficient on-board memory at throughput that can be theoretically adjusted to over $90\%$ of the throughput of a device with sufficient memory.

\subsection{L2L Contribution}
\begin{figure}[ht]
\vskip 0.2in
\begin{center}
\centerline{\includegraphics[width=0.8\columnwidth]{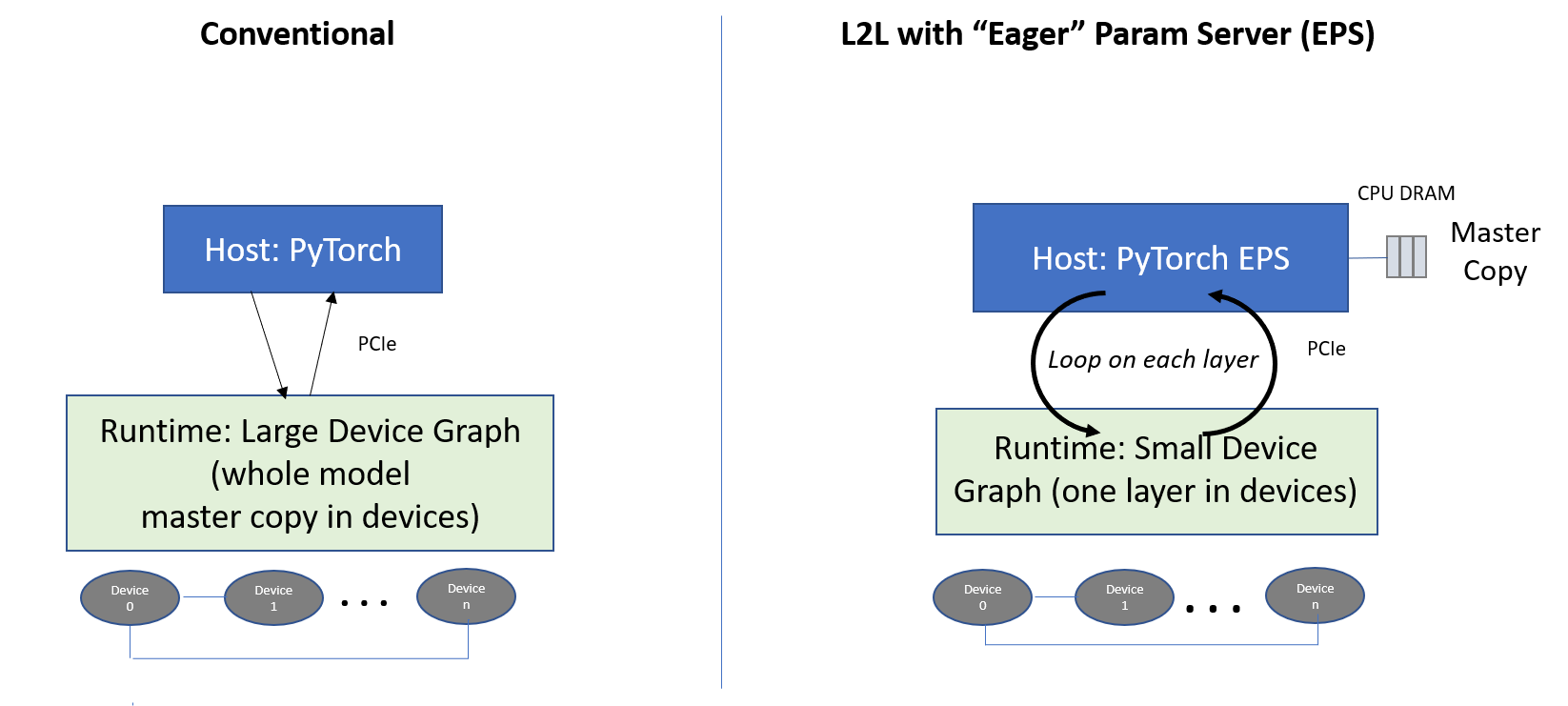}}
\caption{L2L execution with EPS compared with conventional execution.}
\label{fig:eps}
\end{center}
\vskip -0.2in
\end{figure}

In this paper, we propose a new relay-style execution algorithm called L2L (layer-to-layer) that runs models of high weight/activation ratio on a single device. The followings are the main contributions of L2L:
\begin{itemize}
    \item L2L keeps only the executing layer and transit buffers on the device which results in a less than $1GB$ graph on device. The whole model and the optimizer with its state are in the host which relays the next layer through the host-to-device interface after each layer-level iteration on the device. Fig~\ref{fig:eps} shows L2L structure vs conventional approach.
    \item In L2L, a new innerlooping approach is proposed where we run multiple micro-batches (of a minibatch) on one layer at a time. This increases throughput with reduced communication, and it  also enables running a larger batch size on a single device.
    \item Combining the layer-to-layer execution and micro-batching techniques, L2L can scale the model by depth with constant memory without model partitioning.
    \item L2L enables using new precisions such as cross mixed precision (CMP) presented in this paper more efficiently. 
\end{itemize}
L2L allows a researcher to run a very large model independent of depth on a single device or group of devices with a sufficiently large batch size for convergence. 

With L2L, we show that we not only can run BERT-large with higher batch size, less memory and comparable performance than baseline, we demonstrate how L2L runs a gigantic $384$ layer BERT on a single GPU with only $3.69GB$. Every other technique results in out of memory even with $36$ layers. Furthermore, L2L allows fitting models with $50 billion$ parameters on a single $16GB$ V100. 

In theory, L2L can run on top of any model parallelism (pipelined or just partitioned) or checkpointing, so it is complimentary. In particular, it can be combined with DeepSpeed and Zero as the same model memory partitioning can be applied in the eager param-server as each executing device only carries a much smaller part of the model.

\section{Proposed Algorithm}
In this section, we propose a new layer-to-layer (L2L) execution algorithm for running large transformer-based models with constant memory. L2L achieves this by reducing the graph size and layer-wise microbatch looping explained below.
\subsection{L2L Graph Reduction}
In conventional methods, the whole model graph resides on device. In transformer-based models, all encoder layers have similar architectures. Keeping all of the encoders on device is one of the major limitations of running very large models. L2L moves the whole graph to the host which is a special form of param-server we call Eager Param-Server (EPS). 

\begin{figure}[ht]
\vskip 0.2in
\begin{center}
\centerline{\includegraphics[width=0.8\columnwidth]{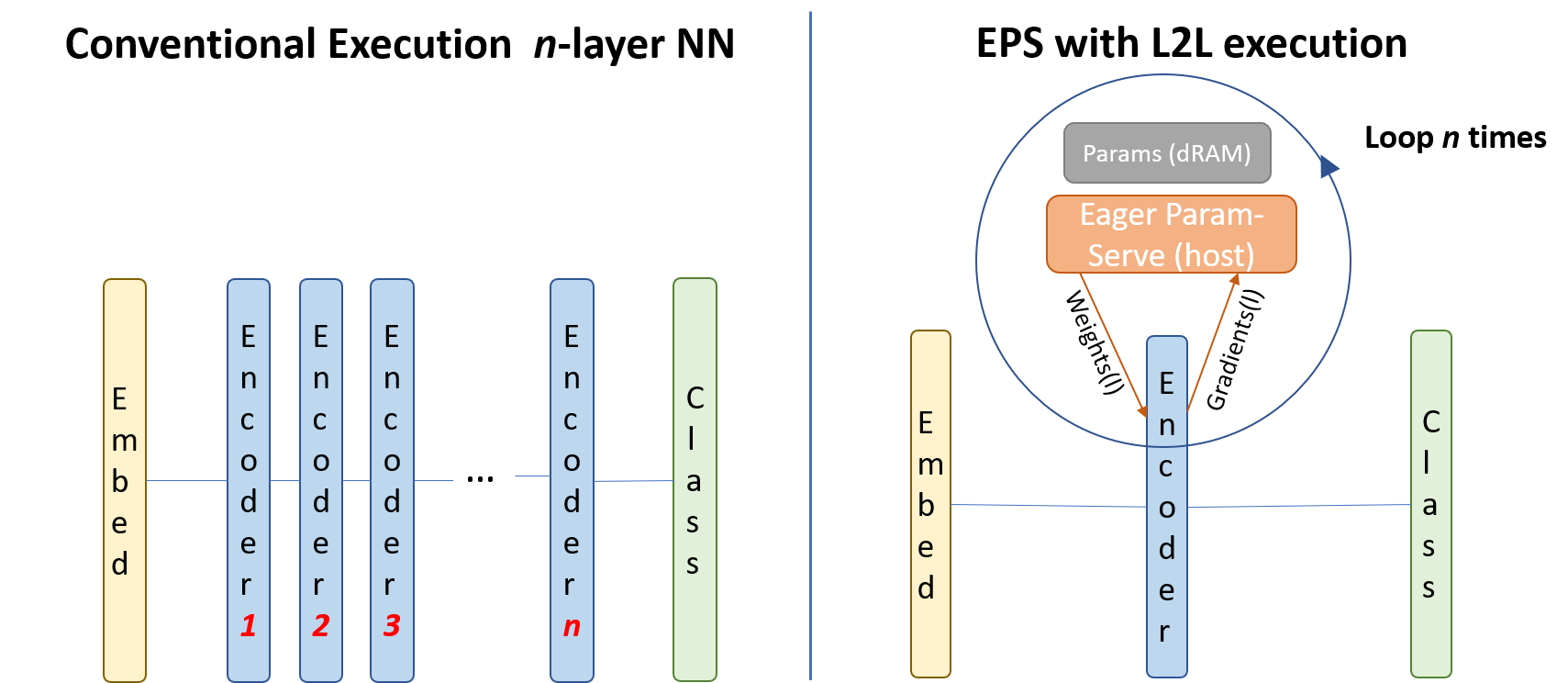}}
\caption{Model graph residing in device for L2L compared with conventional execution.}
\label{fig:l2l}
\end{center}
\vskip -0.2in
\end{figure}

A traditional syncrhonous param-server hosts a coherent space where devices keep their parameters as a state dictionary from which they push all the gradients and update the models at every sync. The EPS - on the other hand – not only services the state space on every layer-level sync, but it also reduces in parallel $eagerly$ which means as soon as the layer-level gradients arrive and in parallel to execution.
Fig~\ref{fig:l2l} shows the execution strategy of L2L for running transformers compared with the conventional approach. 
L2L only keeps one encoder on device which is a flexible encoder layer meaning that it receives each layer's weights from EPS one at a time in forward pass. When running forward pass, only final activations are stashed and all intermediate layer activations are dismissed. In the backward pass, the stashed activations are used one at a time to recompute the forward layer and used for backpropagation of each layer. The gradients computed for each layer will be sent to the EPS and erased from the device memory before going to the next layer. EPS updates weights in the host after receiving all the gradients.

Using this approach, we only need a $3$ layer model to be on device which results in massive saving in on-device memory. In addition, transformer activations can be moved to the EPS if required for further memory saving.

\subsection{Inner Looping in L2L}
\label{sec:innerlooping}
\begin{figure*}[!h]
\begin{subfigure}{1.0\textwidth}
  \centering
  \includegraphics[width=.92\linewidth]{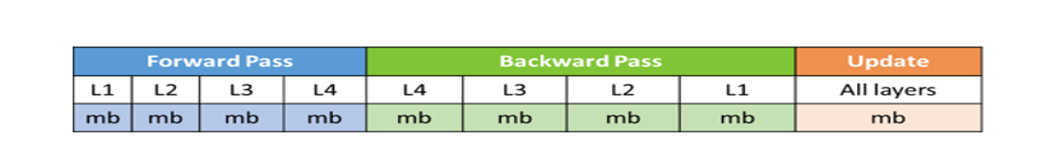}  
  \caption{Conventional data parallel execution}
  \label{fig:conventional}
\end{subfigure}
\begin{subfigure}{1.0\textwidth}
  \centering
  \includegraphics[width=.9\linewidth]{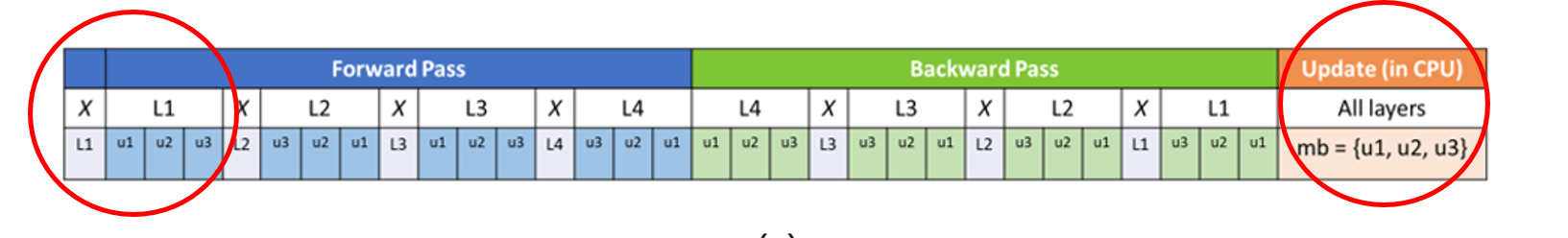}  
  \caption{L2L Execution with micro-batch looping}
  \label{fig:execution}
\end{subfigure}
\caption{L2L execution with micro-batch looping compared with the conventional data parallelism.}
\label{fig:l2l_exec}
\end{figure*}

In conventional data-parallel minibatch execution, each device executes a minibatch of size $mb$ through forward and backward of the whole model and gradient reduction and weight updates are performed afterward. Fig.~\ref{fig:conventional} shows the conventional execution commonly used in the field.
Recompute approach and weights and gradient transfers used in L2L is a loss in effective throughput. But unlike other techniques, L2L can compensate it by keeping the layer on device long enough, i. e. running more data on each layer. To achieve this, L2L proposes a micro-batch looping technique shown in Fig.~\ref{fig:l2l_exec}. The idea is to run a long minibatch $mb$  - if necessary dividing it into a number of microbatches {u1, u2, u3} – on just one layer at a time so that the overall communication overhead of transmitting the layers over a slow host-to-worker interface is  insignificant. Note that increasing the number of microbatches per minibatch is not necessary after the overhead is minimized.  

To give a better picture of the proposed L2L micro-batch looping, we compare this algorithm with the conventional baseline with gradient accumulation. Algorithm~\ref{alg:baseline_ag} shows the execution order in the baseline with accumulated gradients and algorithms~\ref{alg:l2l_s}  shows the execution of the L2L approaches. The main idea here is that L2L inverts the minibatch loop and layer loop. That is the key principle for depth-independent memory sizing.

\begin{minipage}{0.48\textwidth}
\begin{algorithm}[H]
   \caption{Baseline with AG Execution}
   \label{alg:baseline_ag}
\begin{algorithmic}
   \STATE {\bfseries Input:} data $x$, \#layers $layers$, \#uBatches $ub$
   \FOR{$batch$ {\bfseries in} $data$}
   \FOR{$u$ {\bfseries in} $range(ub)$}
   \FOR{$l$ {\bfseries in} $layers$}
   \STATE $act_l = forward(act_{l-1})$
   \ENDFOR
   \FOR{$l$ {\bfseries in} $reverse(layers)$}
   \STATE $g_l += backward(g_{l+1})$
   \ENDFOR
   \STATE $loss = forward(batch)/u$
   \STATE $grads = backward(loss)$
   \STATE $acc\_grads = acc\_grads + grads$
   \ENDFOR
   \FOR{$l$ {\bfseries in} $layers$}
   \STATE $w_l = optimizer(w_l, g_l)$
   \ENDFOR
   \ENDFOR
\end{algorithmic}
\end{algorithm}
\end{minipage}
\hfill
\begin{minipage}{0.48\textwidth}
\begin{algorithm}[H]
   \caption{L2L Execution}
   \label{alg:l2l_s}
\begin{algorithmic}
   \STATE {\bfseries Input:} data $data$,  \#layers $layers$, \#uBatches $ub$
   \FOR{$batch$ {\bfseries in} $data$}
   \FOR{$l$ {\bfseries in} $layers$}
   \FOR{$u$ {\bfseries in} $range(ub)$}
   \STATE $act_l = forward(act_{l-1})$
   \ENDFOR
   \ENDFOR
   \FOR{$l$ {\bfseries in} $reverse(layers)$}
   \FOR{$u$ {\bfseries in} $range(ub)$}
   \STATE $g_l = backward(g_{l+1})$
   \ENDFOR
   \ENDFOR
   \FOR{$l$ {\bfseries in} $layers$}
   \STATE $w_l = optimizer(w_l, g_l)$
   \ENDFOR
   \ENDFOR
\end{algorithmic}
\end{algorithm}
\end{minipage}

\subsubsection{Computational Complexity Analysis on Minimizing Transfer Overhead}

To evaluate the computational complexity of the forward pass, let us assume that the best effective TFLOPs when running a layer is $F$ achieved with a micro-batch size of $ub$, the number of layers is $N$, $L$ is the layer size in $MB$. Also, the PCIe bandwidth from host to device is $B ~GB/sec$ and $c$ is the number of giga operations in a forward pass for $ub$ samples. Without loss of generality, we assume a backward pass which is twice as long as forward, i. e. $2\times c$.
Considering the above assumptions, the transfer time over PCIe is $X = \frac{L}{B}$ in $mSec$. The forward computation for one layer of size $ub$ is $C=\frac{c}{F}$ in $mSec$. The total forward pass time for $ub$ is $N\times (C+X)$. With the backward pass and recompute, the total time for $ub$ can be calculated using Eq.~\ref{eq:time},
\begin{equation}
\label{eq:time}
    \begin{split}
    Total~time & = N\times (C+X) + N\times (3\times C + X) \\
    & = N\times (4\times C + 2\times X),
    \end{split}
\end{equation}
and the throughputs for forward pass and the whole training can be calculated using Eqs.~\ref{eq:forward} and \ref{eq:total}.
\begin{equation}
\label{eq:forward}
    T\_forward = 1000\times \frac{ub}{N\times (C+X)} (samples/sec),
\end{equation}
\begin{equation}
\label{eq:total}
    T\_training = 1000\times \frac{ub}{4\times C+2\times X} (samples/sec).
\end{equation}

With innnerlooping, the overhead of $\frac{X}{C}$ can be vastly reduced. Assuming that $u$ is the number of micro-batches in the inner-loop, forward computation for one layer of size $u\times ub$ is $C' = u\times C$ and the total forward pass time for $ub$ is $N\times (u\times C+X)$. With backward pass and recompute, totoal time for $ub$ can be calculated using Eq.~\ref{eq:time_inner},
\begin{equation}
\label{eq:time_inner}
    \begin{split}
    Total~time & = N\times (u\times C+X) + N\times (3\times u\times C + X) \\
    & = N\times (4\times u\times C + 2\times X),
    \end{split}
\end{equation}
and throughputs for the forward pass and total training can be calculated using Eqs.~\ref{eq:forward_inner} and \ref{eq:total_inner},
\begin{equation}
\label{eq:forward_inner}
    T\_forward = 1000\times \frac{u\times ub}{N\times (C+X)} (samples/sec),
\end{equation}
\begin{equation}
\label{eq:total_inner}
    T\_training = 1000\times \frac{u\times ub}{4\times u\times C+2\times X} (samples/sec).
\end{equation}

The effect of $X$ on the throughput can be diminished by choosing a large size of $u$. For instance, in the case when $X$ is same as $C$, then transfer overhead can be reduced to less than $10\%$ by choosing $u=10$.

\subsection{Cross Mixed Precision}
To reduce the memory requirement, faster kernel computes with less exchanges and getting higher peak TFLOPs, it is necessary to run the models in FP16. However, key NLP workloads don't converge on pure FP16. Nvidia's popular automatic mixed precision (AMP) package provides options to run models with mixed precision which requires keeping the FP32 master copy in the GPU memory.
As explained above, L2L keeps the master copy of the model in the EPS, allowing us to use a new way of running mixed precision called cross mixed precision (CMP). In CMP, we keep an FP32 master copy of the model in the host and run the reduced graph model on GPU with FP16. Using this approach, the optimizer which is running in EPS is updating weights with FP32 precision and the forward and backward pass are in pure FP16. CMP gives us better performance compared to AMP (O2) due to the flexibility L2L provides for running mixed precision.

\section{Experimental Results}

\label{results}
In this section, we present the experimental results for the L2L approach compared with the baseline. 

\subsection{Experimental Data and Setups}
We have used the GLUE dataset~\citep{glue} in our experiments which includes 8 sequence classification tasks. Our experiments are performed on a machine with 4 $16GB$ V100 GPUs and a CPU with $512GB$ memory. The HuggingFace library~\citep{HuggingFaces} is used as a baseline for development and experiments. The pretrained model provided by BERT~\citep{bert}  is used as initial weights for fine-tuning the sequence classification task in both baseline and L2L methods. Table~\ref{config} shows the BERT configuration for both baseline and L2L.
\begin{table*}[hbt!]
\caption{BERT Configuration.}
\label{config}
\vskip 0.15in
\begin{center}
\begin{small}
\begin{sc}
\begin{tabular}{lr}
\toprule
BERT Configuration for baseline and L2L &  \\
\midrule
$\#$transformer layers    & 24 \\
Hidden size & 1024 \\
Intermediate size & 4096 \\
Max sequence Length & 512 \\
Optimizer & LAMB~\citep{lamb} \\
\bottomrule
\end{tabular}
\end{sc}
\end{small}
\end{center}
\vskip -0.1in
\end{table*}

\subsection{Memory and Throughput Test}
The major goal of the L2L is to improve the speed and reduce the memory for running large language models. To demonstrate these goals, we compare L2L and baseline  on running BERT-Large on $4$ V100 GPUs with $16GB$ memory. Table~\ref{tab:mem} shows the throughput and memory comparison performed on the SST-2 which is one of the GLUE tasks. 
\begin{table*}[hbt]
\caption{Memory and throughput comparison between L2L and baseline-AG on $4$ V100 GPUs.}
\label{tab:mem}
\vskip 0.15in
\begin{center}
\begin{small}
\begin{sc}
\begin{tabular}{lcccccr}
\toprule
Method & Precision & Device & uBatch & Total & Throughput  & Memory  \\
 & & Batch Size & Size & Batch Size & (Sample/Sec) & (GB) \\
\midrule
Baseline  & FP32 & 2 & NA & 256 & 16.52 & 10.51 \\
Baseline    & AMP & 2 & NA & 256 & 26.2 & 9.2 \\
\midrule
L2L & FP32 & 64 & 64 & 256 & 22.5 & 9.45 \\
L2L & CMP & 64 & 64 & 256 & 52.48 & 4.96  \\
\bottomrule
\end{tabular}
\end{sc}
\end{small}
\end{center}
\vskip -0.1in
\end{table*}
L2L BERT-Large runs with $50\%$ less memory and twice the throughput. L2L also allows us  to fit the uBatch size of $64$ on each device without any model partitioning while baseline can only fit a device batch size of $2$. Note that there is no inner-looping on this run. 

\subsubsection{Increasing L2L Throughput with Inner Looping}
One of the main contributions of L2L is the innerlooping method discussed in section~\ref{sec:innerlooping}. By increasing the number of uBatches, we can further improve the throughput of L2L. This would, of course, increase the overall minbatch size. Table~\ref{tab:ub} shows L2L results by increasing the number of ubatches from $1$ to $4$.
\begin{table*}[hbt]
\caption{Memory and throughput test for L2L by increasing \#uBatches over $4$ v100-16GBs.}
\label{tab:ub}
\vskip 0.15in
\begin{center}
\begin{small}
\begin{sc}
\begin{tabular}{lcccr}
\toprule
uBatch & \#uBatches & Device & Throughput  & Memory  \\
 Size & & Batch Size  & (Sample/Sec) & (GB) \\
\midrule
64 & 1 & 64 & 52.48 & 3.69\\
64 & 2 & 128 & 70.97 & 3.89 \\
64 & 4 & 256 & 84.91 & 4.27 \\
\bottomrule
\end{tabular}
\end{sc}
\end{small}
\end{center}
\vskip -0.1in
\end{table*}
As the results show, innerlooping increases the throughput by over $60\%$. 

\subsection{L2L Scaling From Single GPU to Multi-GPU}

One of the key aspects of L2L is its potential for scaling near-linearly even without having high-bandwidth links or high-bandwidth networking. This is partly because it uses one shared model for every system of GPUs, which is the only version that needs to be updated at the end of every iteration. And partly because L2L can perform many actions in the background in multi-threaded EPS.
In this paper, we test L2L on single GPU with microbatch size of 64 and compare its performance across multiple GPUs within one system using Pytorch DDP. 

The system does not have NVLinks, so all reductions are in the host with transfers over PCIe. If the GPUs had NV-links, we would be reducing over NV-links first, and each GPU would use its PCI-e link to send out its portion of the gradients.

Table~\ref{tab:scale} shows the performance of L2L by increasing the number of microbatches when running on a single GPU and $4$ GPUs using CMP precision.

\begin{table*}[hbt!] 
\caption{Performance of L2L running on single GPU vs multi-GPU.}
\label{tab:scale}
\vskip 0.15in
\begin{center}
\begin{small}
\begin{sc}
\begin{tabular}{lcccr}
\toprule
\#GPUs & uBatch & \#uBatches & Device & Throughput \\
  & Size & & Batch Size &  Samples/Sec \\
\midrule
1 & 64 & 1 & 64 & 14.57 \\
1 & 64 & 2 & 128 & 18.6 \\
1 & 64 & 4 & 256 & 22.07 \\
\midrule
4 & 64 & 1 & 64 & 52.48 \\
4 & 64 & 2 & 128 & 70.97 \\
4 & 64 & 4 & 256 & 84.91 \\
\bottomrule
\end{tabular}
\end{sc}
\end{small}
\end{center}
\vskip -0.1in
\end{table*} 

We can see that in both single GPU and multi-GPU runs, the throughput increases by increasing the number of microbatches.  We demonstrate  linear
scaling from single GPU to $4$ GPUs even without NV-Links. With NV-Links, the reduction of gradients and loading of layers is much faster as the number of GPUs increase, and it may realize in super-linear scaling. For going parallel across systems, we propose to use the parallel version of L2L (L2Lp) as described in Section~\ref{sec:future} (extended version) where asynchronous communication and parallel optimization can hide the network latencies.

\subsection{Going Beyond BERT}
L2L aims at removing the barriers of memory requirement for running giant transformer-based models on affordable devices. To show this breakthrough, we tried to fit larger models on $16GB$ V100 GPUs by increasing the depth and width of the model.

\subsubsection{Constant Memory by Depth of the Model}
In this experiment, we kept the size of transformer constant and increased the depth of BERT for L2L and baseline. For L2L, we did separate tests by keeping the activation stash on GPU and moving them to the CPU. Table~\ref{tab:const} shows the memory requirement to fit L2L and baseline on a single GPU. 
\begin{table*}[hbt!]
\caption{Memory comparison between the baseline and L2L.}
\label{tab:const}
\vskip 0.15in
\begin{center}
\begin{small}
\begin{sc}
\begin{tabular}{lccccr}
\toprule
Method & uBatch & Device & \#Layer & \#parameters & Memory \\
 & Size & Batch Size &  &  & (GB) \\
\midrule
Baseline  & 2  & 2 & 24 & 300 Million & 9.23 \\
\textbf{Baseline} & 2   & \textbf{2} & \textbf{48} & 600 Million & \textbf{OOM} \\
\midrule
L2L-stash on GPU & 64 & 64 & 24 & 300 Million & 5.22  \\
\textbf{L2L-stash on GPU} & \textbf{64} & \textbf{64} & \textbf{48} & 600 Million & \textbf{6.76} \\
\textbf{L2L-stash on GPU} &  \textbf{64} & \textbf{64} & \textbf{96} & 1.2 Billion & \textbf{9.83} \\
\midrule
L2L-stash on CPU & 64 & 64 & 24 & 300 Million & 3.69  \\
\textbf{L2L-stash on CPU} & \textbf{64} & \textbf{64} & \textbf{96} & 1.2 Billion & \textbf{3.69} \\
\textbf{L2L-stash on CPU} & \textbf{64} & \textbf{64} & \textbf{384} & 4.8 Billion & \textbf{3.69} \\
\bottomrule
\end{tabular}
\end{sc}
\end{small}
\end{center}
\vskip -0.1in
\end{table*}

According to the experiments, we can increase the depth of the model indefinitely (as long as the CPU memory allows) without increasing the GPU memory requirement or partitioning the model. The constant memory aspect of L2L opens up many opportunities for breaking the boundaries of NLP model development and also reducing the cost of deploying models.

\subsubsection{Memory by Width of the Model}
We can increase the width of the model as long as one transformer layer fits in a single GPU and the whole model fits in the CPU memory. To show the performance of L2L on models larger that BERT, we performed a test on a transformer-based model with  settings used in Turing-NLG~\citep{zero}. The model has $78$ transformer layers and each transformer has a hidden size of $4256$ and maximum sequence length is set to $1024$ which results in a $17 billion$ parameter model. Table~\ref{tab:turing} shows the performance of L2L on this model.
\begin{table*}[hbt!]
\caption{Performance of L2L on Turing-NLG-like model.}
\label{tab:turing}
\vskip 0.15in
\begin{center}
\begin{small}
\begin{sc}
\begin{tabular}{lcccccr}
\toprule
uBatch & \#uBatches & Device & \#Layer & \#parameters & Memory & Throughput \\
 Size & & Batch Size &  &  & (GB) & Samples/Sec \\
\midrule
8 & 16 & 128 & 78 & 17 Billion & 6.68 & 0.58 \\
\bottomrule
\end{tabular}
\end{sc}
\end{small}
\end{center}
\vskip -0.1in
\end{table*} 

Table~\ref{tab:turing} proves that L2L enables fitting giant models on a single device with large batch sizes and without requiring to partition the model. Using a machine with $16GB$ V100s and a CPU with $512GB$ memory, we were able to fit models up to $50 billion$ parameters on a single device. 

\subsection{Accuracy Test}
L2L doesn't change the model architecture and it is almost mathematically equivalent to the original version of it. To confirm this, we performed accuracy test on the GLUE tasks for BERT-Large and compared it with baseline. In addition, we have compared the performance of CMP with Nvidia's mixed precision ran on baseline. To do this comparison, we ran both baseline and L2L for 3 epochs on learning rate set to $10^{-3}$. Table~\ref{tab:acc_tab} shows the test results for both FP32 and mixed precision.

\begin{table*}[hbt!]
\caption{Accuracy comparison of L2L and baseline using gradient accumulation for different GLUE dataset tasks.}
\label{tab:acc_tab}
\vskip 0.15in
\begin{center}
\begin{small}
\begin{sc}
  \begin{tabular}{p{0.1\textwidth}c>{\centering}p{1.5cm}cccccr}
    \toprule
    \multirow{2}{*}{Method} & \multirow{2}{*}{Precision} & \multirow{2}{*}{\parbox{1.5cm}{Device Batch Size}} &
      \multicolumn{6}{c}{Accuracy (\%)} \\ [2.5ex]
      & &  & {QNLI} & {SST-2} & {CoLA} & {STS-B} & {MRPC} & {RTE} \\
      \midrule
    Baseline & FP32 & 2  & 91.32 & 93.46 & 58.08 & 89.0 & 88.58 & 66.06 \\
    L2L & FP32 & 64  & 91.46 & 93.92 & 61.05 & 88.50 & 88.30 & 69.67 \\
    \midrule
    Baseline & AMP & 2  & 91.70 & 93.57 & 58.98 & NA & 87.97 & 67.87 \\
    L2L & CMP & 64  & 91.45 & 94.26 & 61.18 & NA & 88.50 & 72.2 \\
    
    \bottomrule
  \end{tabular}
\end{sc}
\end{small}
\end{center}
\end{table*}
Results demonstrate that both L2L and baseline converge to comparable accuracies within a reasonable error bound. As BERT is very sensitive to hyperparameters, the reported accuracies can be further tuned by changing them.

\subsubsection{Validation Curves}
\begin{figure*}[t]
\begin{subfigure}{0.5\textwidth}
  \centering
  \includegraphics[width=1.\linewidth]{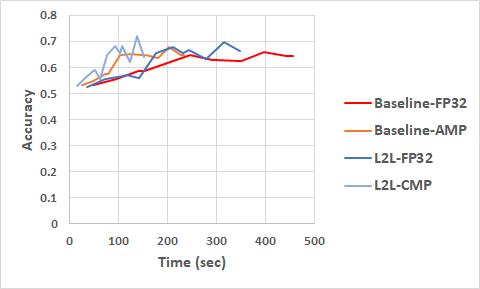}  
  \caption{RTE}
  \label{fig:rte}
\end{subfigure}
\begin{subfigure}{0.5\textwidth}
  \centering
  \includegraphics[width=0.9\linewidth]{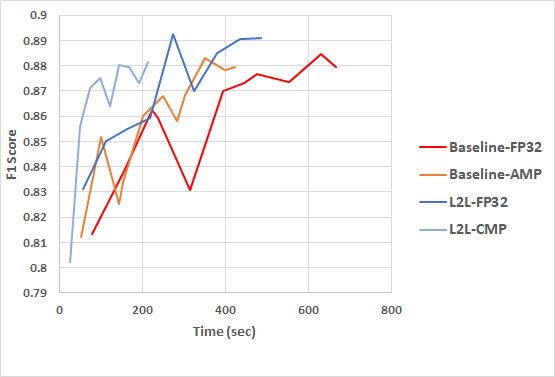}  
  \caption{MRPC}
  \label{fig:mrpc}
\end{subfigure}
\begin{subfigure}{0.5\textwidth}
  \centering
  \includegraphics[width=1.\linewidth]{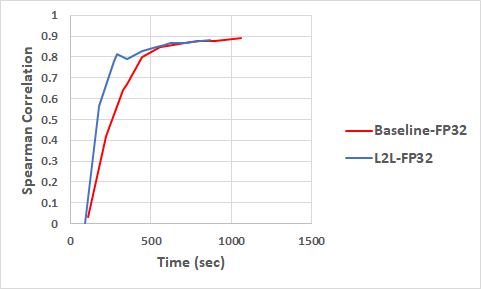}  
  \caption{STS-B}
  \label{fig:mrpc}
\end{subfigure}
\begin{subfigure}{0.5\textwidth}
  \centering
  \includegraphics[width=1.\linewidth]{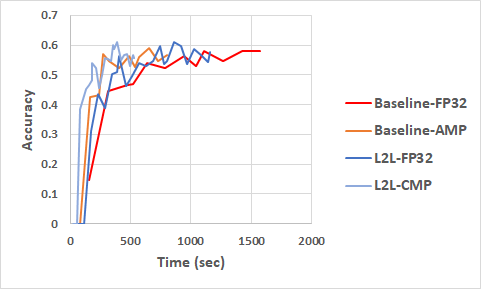}  
  \caption{CoLA}
  \label{fig:mrpc}
\end{subfigure}
\begin{subfigure}{0.5\textwidth}
  \centering
  \includegraphics[width=1.\linewidth]{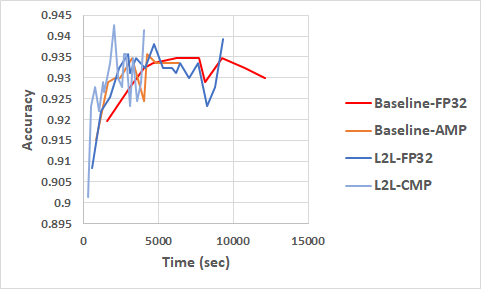}  
  \caption{SST-2}
  \label{fig:mrpc}
\end{subfigure}
\begin{subfigure}{0.5\textwidth}
  \centering
  \includegraphics[width=1.\linewidth]{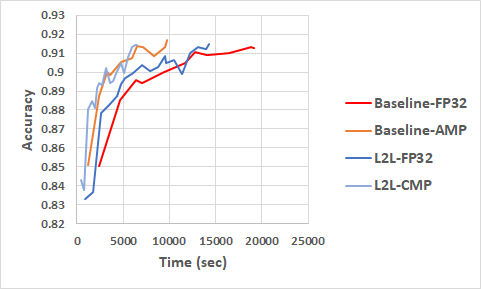}  
  \caption{QNLI}
  \label{fig:mrpc}
\end{subfigure}
\caption{Validation curves for finetuning the GLUE tasks for $3$ epochs.}
\label{fig:validation}
\end{figure*}

 To give a better picture of learning process for both baseline and L2L, we plotted the validation curves for each task when running in both FP32 and CMP precisions. Fig~\ref{fig:validation} shows the validation accuracies when running baseline and L2L for $3$ epochs.

According to the learning curves, we can see that L2L runs faster than baseline and can converge faster in both FP32 and mixed precision.

\section{Conclusion}

Training large language models require massive resources and device memories that are only possible with high-end GPUs and TPUs. 
Moreover, newer ASICs for acceleration are emerging with high FP16 performance and little or no off-chip memory. 
For these reasons, we introduce a new execution paradigm called L2L by elastically using the CPU memory for keeping the model and the optimizer. 
The devices only keep the executing layer while a process in the CPU called EPS prepares and  transmits the next layer. Using inner-looping, L2L reduces the frequency of transfers from EPS to the devices.
EPS also handles reduction and optimization tasks with the potential for virtually linear scaling. An additional benefit was outperforming baseline due to two factors:
(a) faster execution due to relaxed memory, (b) infrequent updates. 

We demonstrate L2L method by running BERT-Large on V100 GPUs with $45\%$ less memory and $40\%$ increase in throughput compared to baseline. 
We also demonstrate that L2L never runs out of memory even when the BERT model grows to 384 layers while all other approaches go out-of-memory. 
In conclusion, the constant-memory nature of this approach allows to scale to arbitrary depth in the number of layers. We enable developers to run very large models on more affordable hardware.
Lastly, each layer can be structurally agnostic to others, encouraging dynamic modeling approaches such as neural architecture search (NAS).

\subsection{Extended Version of L2L (L2Lp) as Future Work}
\label{sec:future}

\begin{figure}[ht]
\vskip 0.2in
\begin{center}
\centerline{\includegraphics[width=0.9\columnwidth]{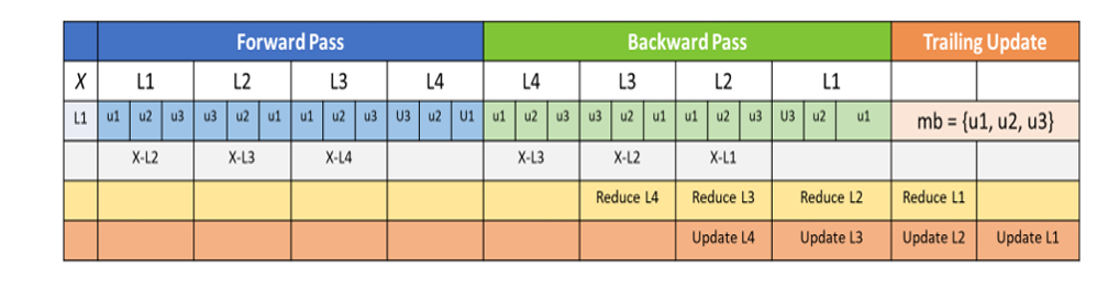}}
\caption{L2Lp's Execution Process.}
\label{fig:l2lp}
\end{center}
\vskip -0.2in
\end{figure}

We introduce a more complete and fully parallel version of L2L called L2Lp. In this version, the EPS performs reduce and weight update in parallel with compute. For example, when layer $l$ completes the backpropagation and sends gradients to the EPS, while it computes the layer $l-1$’s backward, EPS reduces layer $l$ and updates weights for layer $l+1$ in parallel. This is going to further optimizer L2L execution for the best performance. Fig~\ref{fig:l2lp} shows the execution process of L2Lp. 

L2L-p is fully parallel and is projected to scale almost linearly on BERT-Large with virtually zero overhead on thousands of devices, provided batch sizes can be as large as $32K$.  Most of the overhead is simply hidden behind the minibatch size. The only exposed overhead is on the last two layers of reduction and update. This is negligible as the neural network gets deeper for any given model size.

L2L-p does not necessarily require all the bandwidth of the high-speed links (NVLinks) for reduction. It uses a new form of reduce – a “parallel reduce” – where the reduction is wholly in parallel in the EPS for all layers except the last layer which can be through the NVLinks. However, the NV-links will be used for quicker loading of the next layer to offset the slow PCI-e bandwidth across a number of devices. For example, if there are four devices, the EPS will feed each device one-fourth of the weights over PCIe. Then the devices gather the weights over the high-speed NVLinks at full throttle.

\section*{Broader Impact}

By proposing L2L, we are following a path to democratize AI and make NLP models accessible for everyone. L2L allows researchers to train such models with less memory and higher throughput resulting in massive cost saving. It provides opportunities for using affordable devices to train and deploy giant NLP models. In addition, L2L provides flexibility to pursue neural architecture search considering the affordability and dynamic nature of the GPU model.

We hope this new execution paradigm will also influence the hardware industry that is currently investing in single-tier devices with brute-force High Bandwidth Memory technologies and high speed links to also consider a two-tier approach to training where the top tier is responsible for the model and optimization (EPS) while the device tier is responsible for executing the layer.

\section*{Acknowledgements}
This paper and the research behind it would not have been possible without the exceptional support of our manager and colleagues. We would especially like to thank Tiyasa Mitra, Mohit Mittal, Layali Rashid, Marc Tremblay, and Rajiv Kapoor for their advice and support during the development and publishing of this paper.

\bibliographystyle{apalike}
\bibliography{neurips_2020}

\end{document}